\documentclass[journal]{IEEEtran}
\usepackage{lineno,hyperref}
\usepackage{bbm}
\usepackage{bm}
\usepackage{amsfonts}
\usepackage{amsmath}
\usepackage{mathrsfs}
\usepackage{amssymb}
\usepackage{enumerate}
\usepackage{graphicx}
\usepackage{subfigure}
\usepackage{booktabs}
\usepackage{graphicx}
\usepackage{setspace}
\usepackage{algorithm}
\usepackage{algorithmic}
\usepackage{setspace}
\usepackage{multirow}
\usepackage{array}
\usepackage{color}
\usepackage{cite}
\usepackage{geometry}
\ifCLASSINFOpdf
\else
\fi
\begin{document}
\title{CLIP-Driven Fine-grained Text-Image Person Re-identification}
\author{Shuanglin~Yan, Neng~Dong, Liyan~Zhang, Jinhui~Tang, \emph{Senior Member, IEEE}
\thanks{S. Yan, N. Dong, and J. Tang are with the School of Computer Science and Engineering, Nanjing University of Science and Technology, Nanjing 210094, China  (e-mail: shuanglinyan@njust.edu.cn; neng.dong@njust.edu.cn; jinhuitang@njust.edu.cn).}
\thanks{L. Zhang is with the College of Computer Science and Technology, Nanjing University of Aeronautics and Astronautics, Nanjing 210016, China (e-mail: zhangliyan@nuaa.edu.cn).}
\thanks{Corresponding author: Liyan~Zhang.}
}
\markboth{Journal of \LaTeX\ Class Files}%
{Shell \MakeLowercase{\textit{et al.}}: Bare Demo of IEEEtran.cls for IEEE Journals}
\maketitle
\begin{abstract}
Text-Image Person Re-identification aims to retrieve the image corresponding to the given text query from a pool of candidate images. Existing methods employ prior knowledge from single-modality pre-training to facilitate learning, but lack multi-modal correspondence information. Besides, due to the substantial gap between modalities, existing methods embed the original modal features into the same latent space for cross-modal alignment. However, feature embedding may lead to intra-modal information distortion. Recently, Contrastive Language-Image Pretraining (CLIP) has attracted extensive attention from researchers due to its powerful semantic concept learning capacity and rich multi-modal knowledge, which can help us solve the above problems. Accordingly, in this paper, we propose a CLIP-driven Fine-grained information excavation framework (CFine) to fully utilize the powerful knowledge of CLIP for TIReID. To transfer the multi-modal knowledge effectively, we perform fine-grained information excavation to mine intra-modal discriminative clues and inter-modal correspondences. Specifically, we first design a multi-grained global feature learning (MGF) module to fully mine the discriminative local information within each modality, which can emphasize identity-related discriminative clues by enhancing the interactions between global image (text) and informative local patches (words). MGF can generate a set of multi-grained global features for later inference. Secondly, cross-grained feature refinement (CFR) and fine-grained correspondence discovery (FCD) modules are proposed to establish the cross-grained and fine-grained interactions (image-word, sentence-patch, word-patch) between modalities, which can filter out unimportant and non-modality-shared image patches/words and mine cross-modal correspondences from coarse to fine. CFR and FCD are removed during inference to save computational costs. Note that the above process is performed in the original modality space without further feature embedding. Extensive experiments on multiple benchmarks demonstrate the superior performance of our method on TIReID.

\end{abstract}
\IEEEpeerreviewmaketitle
\section{Introduction}
\IEEEPARstart{P}{erson} Re-identification (ReID) is a popular and challenging task in computer vision. In the past decade, ReID has made remarkable progress~\cite{pcb, AIESL, pifhsa, GLMC}, and has been successfully applied in some practical scenarios. Most existing ReID approaches assume that the pedestrian’s images can be captured across disjoint cameras, and tend to ignore the situation that pedestrian images cannot be obtained in some complex or special scenes, such as some remote roads without cameras or where pedestrians are completely occluded. Although pedestrian images are not available, we can find some witnesses at the scene and search for the target pedestrian by the witness's language description, that is, text-image person re-identification (TIReID)~\cite{GNA}. Due to its great practical value, TIReID has attracted increasing attention from both academia and industry.

\begin{figure}[t!]
    \centering
    \includegraphics[height=4.0in,width=3.4in,angle=0]{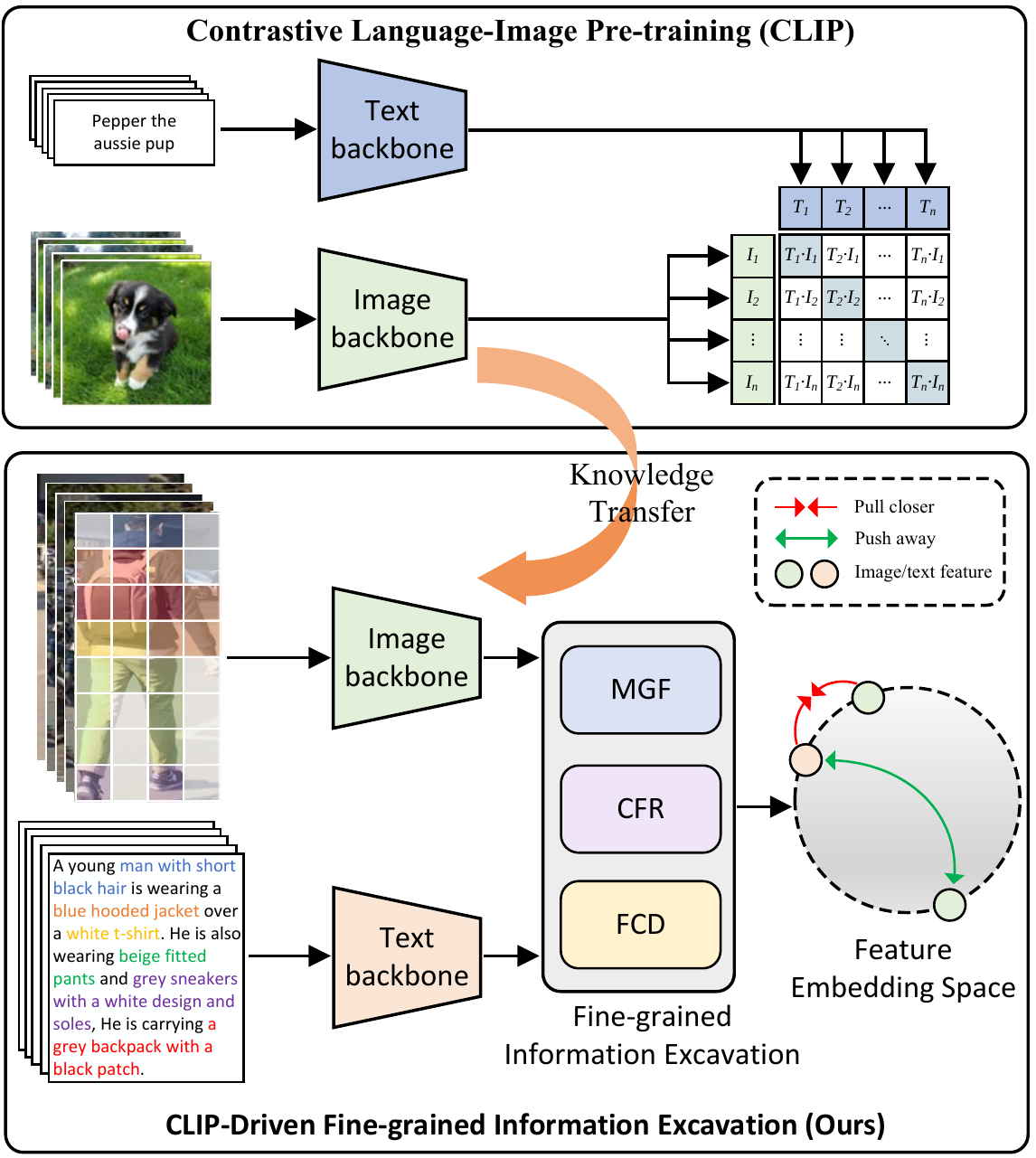}
    \caption{The Motivation for our proposed method. (a) CLIP learns visual representation with natural language supervision using web-scale image-text data, and the learned visual representation contains rich semantic information and cross-modal correspondence information. (b) We explore leveraging the powerful knowledge of CLIP for TIReID. CLIP is trained to focus only on instance-level representation, while TIReID requires fine-grained discriminative clues. To take full advantage of rich prior knowledge from CLIP, our CFine proposes three innovative modules (MGF, CFR, and FCD) to mine intra-modal discriminative clues and inter-modal fine-grained correspondences.}
    \label{Fig:1}
\end{figure}

As a fine-grained cross-modal retrieval task, the key of TIReID is to mine the fine-grained information of images and texts, and establish their correspondences. In recent years, many effective methods~\cite{vitaa, CMAAM, NAFS, DSSL, SSAN, tipcb} have been proposed, all of which follow the same structural design of “image/text backbone + feature embedding”, where image/text backbone first extract image/text features, and then feature embedding (method-specific) embeds the extracted image and text features into a joint space for cross-modal alignment. Image/text backbone generally utilizes external knowledge to facilitate learning, while mainly initializes backbone by single-modality pre-training (e.g., the pre-trained ResNet~\cite{resnet} and ViT~\cite{vit} on ImageNet, the pre-training language model BERT~\cite{bert}), which \textbf{lacks multi-modal correspondence information}. In addition, existing methods generally believe that feature embedding is the core module of the TIReID task, which is crucial for narrowing the semantic gap between modalities. However, several works~\cite{lat, hge} have shown that projecting different modalities into the joint space may lead to \textbf{intra-modal information distortion} due to the distinct data properties between images and texts. So we need to reconsider whether feature embedding is necessary for the TIReID task.

Recently, the remarkable success of visual-language pre-training (VLP) has shown its ability to learn semantically rich and high-quality visual concepts with natural language supervision, and the most representative work is Contrastive Language-Image Pre-training (CLIP)~\cite{clip}. Compared with single-modal pre-training, CLIP contains abundant multi-modal knowledge. In light of the power of CLIP, some recent works have attempted to exploit its ample knowledge to various tasks and achieved impressive results, such as video-text retrieval~\cite{clip4clip}, referring image segmentation~\cite{cris}, dense prediction~\cite{denseclip}, video understanding~\cite{gvr}. Besides, the semantic-level visual concept representation capacity of CLIP makes it possible to align image and text in the original modality space.  Inspired by this, we explore how to transfer the CLIP model to TIReID in this paper. We found that fine-tuning CLIP directly on TIReID is effective. But CLIP is trained to only pay attention to instance-level representation (image-level, sentence-level), while TIReID requires the model to focus on fine-grained information and inter-modal correspondences to distinguish the sutble differences between pedestrians (as Figure~\ref{Fig:1}). Thus, the direct usage of CLIP can be sub-optimal for TIReID due to the task gap. 

To fully exploit the powerful knowledge of the CLIP model, we propose a novel \textbf{C}LIP-driven \textbf{F}ine-grained \textbf{in}formation \textbf{e}xcavation framework, namely \textbf{CFine}, for TIReID. As shown in Figure \ref{Fig:1}, CFine mainly includes two parts: modality-specific feature extraction and fine-grained information excavation. To be specific, CFine first adopts modality-specific encoders to extract the image and text representations, and then the fine-grained information excavation part is exploited to mine intra-modal discriminative details and inter-modal fine-grained correspondences for better transferring knowledge of CLIP to TIReID. The fine-grained information excavation includes three main components. First, we propose a multi-grained global feature learning (MGF) module to fully mine the identity-related sutble clues within each modality. In this module, we first design a token selection process to pick out a set of informative tokens (discriminative patches/words) based on the self-attention score between class token and local tokens for each modality. Then, the informative token set is split into multiple subsets and is fed into a global-local decoder (GLD) to generate a set of multi-grained global features by enhancing the interactions between global image (text) and local discriminative patches (words). Second, we design a cross-grained feature refinement (CFR) module to filter out non-modality-shared information in the selected tokens by computing cross-grained similarities (image-word, sentence-patch) between modalities, and establish the rough cross-modal correspondence. Third, to establish inter-modal fine-grained correspondence, we propose a fine-grained correspondence discovery (FCD) module to discover the relationship between words and image patches. Note that the entire learning process of CFine is performed in the original feature space without further feature embedding, and optimized in an end-to-end manner. During inference, CFR and FCD are removed, and multi-grained global image and text features generated by MGF are used for cross-modal retrieval. Our main contributions are summarized as follows:
\begin{itemize}
\item We propose a CLIP-driven fine-grain information excavation framework to transfer the knowledge of CLIP to TIReID, achieving fine-grained text-image alignment without further feature embedding. To our best knowledge, we are the first to leverage the ample cross-modal knowledge from VLP to facilitate learning for TIReID.

\item We take full advantage of rich multi-modal knowledge from CLIP via three innovative modules, i.e., multi-grained global feature learning, cross-grained feature refinement, and fine-grained correspondence discovery.

\item We conduct extensive experiments on three benchmarks to validate the effectiveness of our CFine. CFine performs significantly better than previous methods and reaches 69.57 Rank-1, 60.83 Rank-1, 50.55 Rank-1 on the CUHK-PEDES, ICFG-PEDES, and RSTPReid, respectively, which outperforms the previous SOTA method by +5.13\%, +4.79\%, and +3.85\%.
\end{itemize}

The remainder of the paper is organized as follows. We first review the related works in Section \uppercase\expandafter{\romannumeral2}; Section \uppercase\expandafter{\romannumeral3} describes the proposed CFine in detail; Section \uppercase\expandafter{\romannumeral4} reports extensive experimental results and analysis; and finally the paper is summarized in Section \uppercase\expandafter{\romannumeral5}.

\section{Related work}
\subsection{Text-Image Person Re-identification}
TIReID is a class of multi-modal tasks~\cite{qda, mirrorgan}, which was first proposed by \cite{GNA}. Compare with general cross-modal retrieval tasks, TIReID is more challenging as its fine-grained property. The key to TIReID is cross-modal alignment, existing methods can be broadly classified into two classes according to the alignment strategy: cross-modal interaction-based and cross-modal interaction-free methods. Cross-modal interaction-based methods~\cite{GNA, HGAN, MIA, PMA, SCAN, NAFS, mgel, DSSL} focus on mining local correspondences (e.g., patch-word, patch-phrase) between images and texts by using attention mechanism to predict the matching score for image-text pairs. Gao \emph{et al}.~\cite{NAFS} designed a contextual non-local attention, which adaptively aligns image and text features across all scales in a coarse-to-fine way according to their semantics. The cross-modal interaction mechanism is a double-edged sword, and its advantage lies in better aligning image-text pairs and reducing modality gap through sufficient cross-modal interaction, so such methods can achieve superior performance. But its disadvantage is the high computational cost, which greatly reduces the practicability of such methods.

For cross-modal interaction-free methods, some early works~\cite{Dual, CMPM, MCCL, A-GANet, TIMAM, CMKA, vitaa, CMAAM} mainly focused on designing network and optimization loss to learn aligned image and text embeddings in a joint latent space. Zhang \emph{et al}.~\cite{CMPM} designed a cross-modal projection matching (CMPM) loss and a cross-modal projection classification (CMPC) loss to learn discriminative image-text embeddings, which is one of the most commonly used cross-modal matching losses for TIReID. These early methods are efficient but their performance is not satisfactory. In recent years, some effective and lightweight models~\cite{SSAN, tipcb, LapsCore} have been proposed, which can achieve better performance than the first type of methods without complex cross-modal interaction. With the success of Transformer in various visual and language tasks, several Transformer-based methods~\cite{saf, lgur} have been proposed recently and achieved state-of-the-art performance. All in all, TIReID has achieved remarkable progress over the past few years.

However, existing methods initialize the network through single-modality pre-trained model parameters, ignoring the multi-modal correspondence information. Besides, the image and text embeddings extracted from the above-initialized network over-mine the information within a single modality, which increases the difficulty of cross-modal alignment and network optimization. Recently, visual-language pre-training (VLP) has attracted growing attention, especially CLIP~\cite{clip}, whose own advantages can effectively alleviate the above problems. Thus, we explore leveraging the powerful knowledge of the CLIP model for TIReID in the paper.

\subsection{Vision-Language Pre-Training}
The paradigm of "pre-training and fine-tuning" is one of the most important paradigms to drive the development of computer vision community. Its basic process is that the model is initialized by pre-trained model parameters on large-scale datasets, and then fine-tuned on various downstream tasks. In this paradigm, the quality of the pre-training model plays a vital role in the optimization difficulty and performance of the model during fine-tuning. In the past decade, the pre-training model in single modal domain~\cite{vit, resnet, bert} has achieved great success. Recently, many works~\cite{clip, align, filip, locvtp, visualbert, uniter} have attempted to extend the pre-training model to multi-modal field, that is, visual-language pre-training (VLP), and made remarkable progress. The mainstream VLP models can be divided into two categories according to the pre-training tasks: (1) image-text contrastive learning tasks, which align images and texts into a shared space through cross-modal contrastive learning loss, e.g., CLIP~\cite{clip}, ALIGN~\cite{align}, FILIP~\cite{filip}; (2) language modeling based task, some auxiliary tasks (Masked Language/Region Modeling, image captioning, text-grounding image generation) are used to establish the correspondences between images and texts, e.g., VisualBERT~\cite{visualbert}, UNITER~\cite{uniter}. The VLP models can bring greater performance gains for cross-modal tasks and fine-grained visual tasks, which is confirmed by a large number of follow-up works. We also expect to leverage ample multi-modal knowledge of VLP to further advance the TIReID task.

\subsection{CLIP-Based Fine-tuning}
As one of the most representative VLP models, Contrastive Language-Image Pre-training (CLIP)~\cite{clip} has attracted much attention. Different from the traditional image-based supervised pre-training model, CLIP employs natural language to supervise the learning of visual features with contrastive learning on web-scale image-text data. Benefiting from semantic-level language supervision, the visual network can learn high-quality visual features with rich semantic information, which has an impressive positive impact on cross-modal tasks and fine-grained visual tasks. Recently, a lot of follow-ups~\cite{clip4clip, clip2video, x-clip, centerclip, clip4caption, cris, gvr, blockmix, pointclip, denseclip} has been put forward to fine-tune CLIP to various downstream tasks. The most common is to adapt CLIP to some cross-modal tasks, such as video-text retrieval~\cite{clip4clip, clip2video, x-clip, centerclip}, video caption~\cite{clip4caption}, referring image segmentation~\cite{cris}. Besides, some efforts have recognized the semantic-level and high-quality visual concept representation capacity of CLIP, and applied it to some fine-grained visual tasks, including dense prediction~\cite{denseclip}, point cloud understanding~\cite{pointclip}, video recognition~\cite{gvr}, and achieved impressive results. As a cross-modal retrieval task as well as a fine-grained recognition task~\cite{agpf, bsfa}, TIReID can also benefit from CLIP. Thus, in this paper, we try to explore an effective framework to fully transfer the CLIP model to the TIReID task.

\begin{figure*}[t!]
  \centering
  \includegraphics[width=6.6in,height=4.7in]{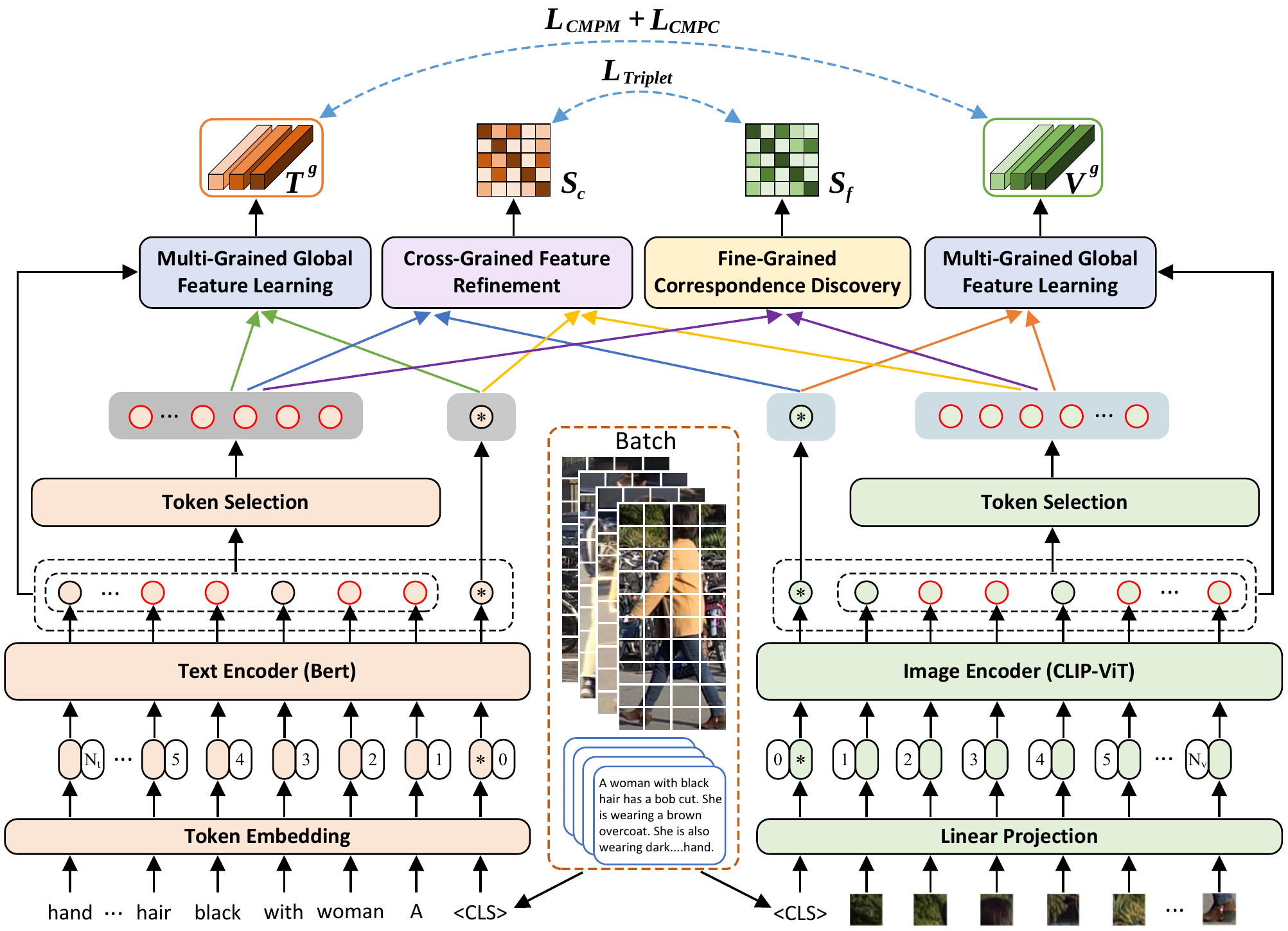}\\
  \caption{Overview of the proposed CFine. Given image-text pairs, we first extract global image/text features and local patch/word features by image/text encoder. After that, the token selection is performed to select some informative patch/word features from local patch/word features. The above global image/text features, local patch/word features, and these selected informative patch/word features are sent to the multi-grained global feature learning module, which can help reinforce the fine-grained clues and generate a set of multi-grained global image/text features $\bm V^g/\bm T^g$. Cross-grained feature refinement and fine-grained correspondence discovery modules take global image/text features and selected informative patch/word features as input to filter out non-modality-shared patches/words and mine cross-modal correspondences from coarse to fine, which can generate two similarity matrices $\bm S_c/\bm S_f$ for a batch of image-text pairs and be removed during inference. Finally, $\bm V^g/\bm T^g$ and $\bm S_c/\bm S_f$ are supervised by CMPM+CMPC loss and Triplet ranking loss for cross-modal alignment, respectively.}
  \label{Fig:2}
\end{figure*}

\section{Methods}
\subsection{Motivation and Overview of CFine}
Existing prior methods in TIReID utilize external knowledge from single-modality pre-training to facilitate learning, which is short of multi-modal correspondence information. However, it is unaffordable to directly learn a visual-language pre-training model for TIReID from scratch due to it requires inaccessible large-scale image-text data and expensive training resources. Recently, visual-language pre-training (VLP) has made significant progress and shown its rich cross-modal correspondence information and powerful visual representation capacity. To leverage the powerful capacity of VLP, several efforts have attempted to transfer the prior knowledge of VLP to various downstream tasks, achieving impressive results. Inspired by these works, our method builds upon the currently most popular VLP model, namely CLIP~\cite{clip}, and extends it with fine-grained information excavation to better adapt the TIReID task, instead of learning a new pre-training model from scratch. An overview of the proposed CFine is illustrated in Figure \ref{Fig:2}. 

Given a set of pedestrian images $\mathcal{V}$ and text descriptions $\mathcal{T}$, we first feed them to the dual encoders to extract the image and text features. Second, to better adapt to TIReID, the fine-grained information excavation is performed to mine intra-modal fine-grained information and inter-modal fine-grained correspondences. For fine-grained information excavation, three modules are proposed: (1) The multi-grained global feature learning (MGF) mines discriminative local clues according to informative tokens at different levels, and generates a set of multi-grained features; (2) A cross-grained feature refinement (CFR) is proposed to filter out unnecessary information in image/text and ensure the confidence of informative tokens; (3) A fine-grained correspondence discovery (FCD) is proposed to establish local fine-grained correspondences between patches and words. Finally, the similarity between the above-learned image and text representations is computed by the cosine similarity function, whose goal is to maximize the similarity if the image and text are matched, and minimize it otherwise.

\begin{figure*}[t!]
  \centering
  \includegraphics[width=5.8in,height=3.6in]{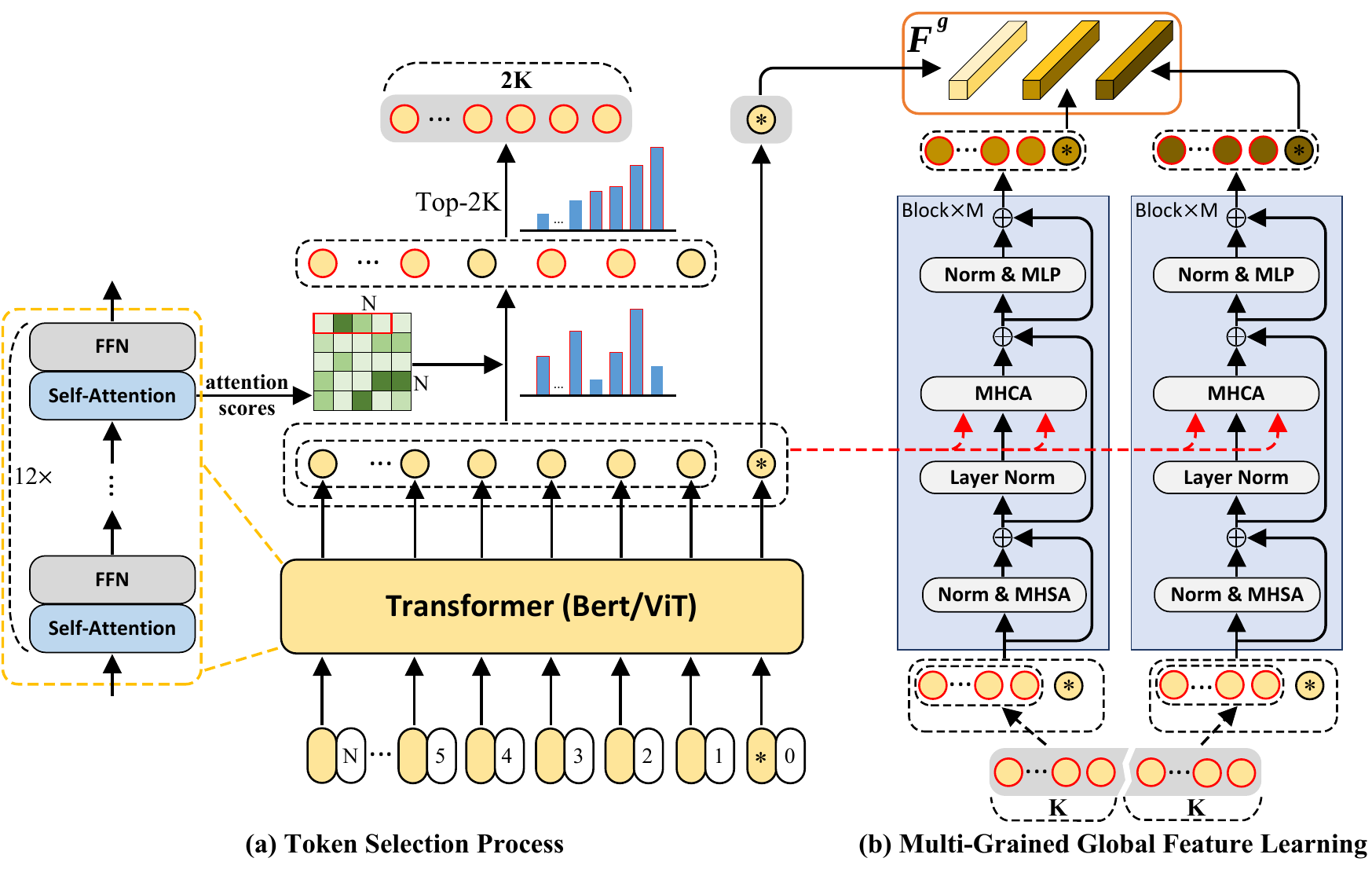}\\
  \caption{(a) Illustration of token selection process and (b) structure of multi-grained global feature learning (MGF) module, where $\bm F^g (\bm V^g/\bm T^g)$ represents the multi-grained global feature set.}
  \label{Fig:3}
\end{figure*}

\subsection{Dual Encoders}
The structure of CLIP is shown in Figure~\ref{Fig:1}, which includes an image encoder and a text encoder, both of which are composed of a feature extractor and a projector. The image and text feature extractors extract features through a ViT with a width of 768 and a Transformer with a width of 512 respectively, while the projectors map image and text features to a 512-dimensional latent space, in which image and text are aligned through the contrastive objective. The most direct way is to fine-tune CLIP on the TIReID dataset. However, several works~\cite{lat, hge} have shown that the projectors may lead to intra-modal information distortion. This is unacceptable for TIReID which relies on fine-grained information, especially images. If the projector is removed, the dimensions of the two cannot be unified. Thus, in the paper, we only use the image encoder of CLIP with the projector removed as our image encoder. For text, we use another pre-training language model BERT~\cite{bert} as text encoder. In addition, in order to make a fair comparison with existing methods, we also use ViT pre-trained on ImageNet~\cite{imagenet} as image encoder.

\textbf{Image Representation.} Given an image $I \in \mathcal{V}$, a visual tokenization process is first performed to convert the image to a discrete token sequence of length $N_{v}$. A learnable [$CLS$] token is attached to the beginning of the sequence as an image-level representation. Finally, the token sequence of length $N_{v}+1$ is fed into the transformer of ViT. The output of image encoder is represented as $\bm V=\{\bm v_{g}, \bm v_{1}, \bm v_{2}, ..., \bm v_{N_{v}}\}\in \mathbb{R}^{(N_{v}+1)\times d}$, where $\bm v_{g}$ is the image-level global feature, $\{\bm v_{1}, \bm v_{2}, ..., \bm v_{N_{v}}\}$ is the patch-level local features.

\textbf{Text Representation.} For a text $T \in \mathcal{T}$, we directly use the pre-trained BERT~\cite{bert} as text encoder to generate text representation. Specifically, the lower-cased byte pair encoding (BPE) with a 30522 vocabulary size is firstly used to tokenize the text $T$. Then, the textual token sequence is padded with [$CLS$] token at the beginning. Finally, the token sequence of length $N_{t}+1$ is fed into the text encoder to generate the sentence-level global feature $\bm t_{g}\in \mathbb{R}^{d}$ and the word-level local features $\{\bm t_{1}, \bm t_{2}, ..., \bm t_{N_{t}}\}\in \mathbb{R}^{N_{t}\times d}$.

Due to the substantial gap between the upstream pre-training task and the downstream TIReID task, $\bm v_{g}$ and $\bm t_{g}$ only involves instance-level modality information and cross-modal correspondences, which is short of fine-grained information that is critical to TIReID. Accordingly, in the following, we conduct fine-grained information excavation (MGF, CFR, and FCD) to fully mine intra-modal discriminative local clues and inter-modal fine-grained correspondences.

\subsection{Multi-Grained Global Feature Learning}
Due to the subtle visual variations among different pedestrians in ReID, it is crucial to fully mine the fine-grained information of images/texts for distinguishing different pedestrians. Most existing TIReID methods mine fine-grained information by learning a set of local features. Unlike them, in this paper, we mine local information at different levels to learn global features at multiple granularities. Benefiting from the global dependency modeling capability of self-attention, Transformer achieves impressive results on various tasks. However, self-attention treats each local token in the same way to calculate the attention weight and then computes a weighted sum of all local tokens to generate a global feature. The global feature is dominated by all local tokens, and this way of considering all local tokens simultaneously reduces the influence of some important local tokens. Especially for fine-grained recognition tasks, the way will bring serious discriminative information loss. To solve this problem, instead of using all tokens to learn global features, we only select informative tokens to form multiple token sequences, and then send them to the global-local decoder to learn a set of multi-grained global features.

\textbf{Token Selection}. Class token as the output of Transformer is used for classification or recognition, which is obtained by weighted aggregation of all local tokens. The weight reflects the correlation between class token and each local token. The larger the weight, the greater the contribution of this local token to the class token, and the more important it is to the task. Therefore, we select informative tokens based on the correlation between local tokens and class token~\cite{trt, dcal, ts2net}. Specifically, the self-attention of each Transformer block can generate an attention map of size $(1+N)\times (1+N)$, which reflects the correlation among the input $1+N$ tokens (the first is the class token). The first row of the attention map represents the dependency between class token and local tokens. In this paper, we take the attention map $\bm A\in \mathbb{R}^{(1+N)\times (1+N)}$ generated by the self-attention of the last Transformer block, and the correlation score between class token and local tokens is $\bm m= \bm A[0, 1:]\in \mathbb{R}^{N}$. We select the top $2K$ tokens from the $N$ local tokens output by Transformer that corresponds to the top $2K$ highest scores in $\bm m$ to construct a new discriminative local token sequence. The token selection process is shown in Figure~\ref{Fig:3}(a).

We perform the token selection process separately for images and texts. For image $I$, $2K_v$ tokens are chosen from $\{\bm v_{1}, \bm v_{2}, ..., \bm v_{N_{v}}\}$. For text $T$, $2K_t$ tokens are chosen from $\{\bm t_{1}, \bm t_{2}, ..., \bm t_{N_{t}}\}$. The selected image and text token sequences are denoted as $\bm V^{s}=\{\bm v^{s}_{1}, ..., \bm v^{s}_{K_{v}}, \bm v^{s}_{K_{v}+1}, ...,\bm v^{s}_{2K_{v}}\}\in \mathbb{R}^{2K_{v}\times d}$ and $\bm T^{s}=\{\bm t^{s}_{1}, ..., \bm t^{s}_{K_{t}}, \bm t^{s}_{K_{t}+1}, ...,\bm t^{s}_{2K_{t}}\}\in \mathbb{R}^{2K_{t}\times d}$, where $H_{v^{s}_{1}}\textgreater H_{v^{s}_{K_{v}}}$ and $H_{t^{s}_{1}}\textgreater H_{t^{s}_{K_{t}}}$, $H$ denotes the amount of information contained in the token. $K_v=R_vN_v$ and $K_t=R_tN_t$, where $R_v$ and $R_t$ represent the selection ratio of image and text tokens, respectively.

\textbf{Global-Local Decoder}. We design a global-local decoder (GLD) to highlight the discriminant local information in images and texts, and improve the discriminability of global features. As shown in Figure \ref{Fig:3}(b), the GLD containing $M$ blocks generates a set of multi-grained discriminative global features with the above-selected token sequence as the input. The selected token sequence only contains informative tokens, discarding redundant tokens. To fully mine discriminative fine-grained information, we split the selected token sequence into two sub-sequences, which correspond to different discriminant granularities. The former is a high-level discriminant sequence with the top $K$ most informative tokens, while the latter is a middle-level discriminant sequence with the remaining $K$ informative tokens. Two sub-sequences are fed into GLD to highlight different-grained local information. Specifically, taking image as an example, $\bm V^{s}$ is divided into the high-level sequence $\bm V^{s}_h\in \mathbb{R}^{K_{v}\times d}$ and middle-level sequence $\bm V^{s}_m\in \mathbb{R}^{K_{v}\times d}$, which are respectively prepended with a [$CLS_h$] token and a [$CLS_m$] token, are then fed into GLD. In one GLD block, $\bm V^{s}_h$/$\bm V^{s}_m$ is first sent into the multi-head self-attention ($MHSA$) layer to propagate the information of these informative tokens into the class token.
\begin{equation}\
\begin{aligned}
\hat{\bm V}^{s}_h/\hat{\bm V}^{s}_m=MHSA(Norm(\Bar{\bm V}^{s}_h/\Bar{\bm V}^{s}_m))+\Bar{\bm V}^{s}_h/\Bar{\bm V}^{s}_m
\end{aligned}
\end{equation}
where $\Bar{\bm V}^{s}_h/\Bar{\bm V}^{s}_m=[\bm v_{CLS_h}, \bm V^{s}_h]/[\bm v_{CLS_m}, \bm V^{s}_m]$, $Norm(\cdot)$ denotes Layer Normalization. After that, $\hat{\bm V}^{s}_h$ and $\hat{\bm V}^{s}_m$ are fed into the multi-head cross-attention ($MHCA$) layer to compute the cross-attention between $\hat{\bm V}^{s}_h/\hat{\bm V}^{s}_m$ and $\bm V$, which can highlight not only the informative tokens themselves but also other associated contextual information. Finally, the output of $MHCA$ is further fed into the multi-layer perceptron ($MLP$) layer to generate multiple discriminative global features at different granularities.
\begin{equation}\
\begin{aligned}
~~~~\Tilde{\bm V}^{s}_h/\Tilde{\bm V}^{s}_m=MHCA(Norm(\hat{\bm V}^{s}_h/\hat{\bm V}^{s}_m), \bm V)+\hat{\bm V}^{s}_h/\hat{\bm V}^{s}_m
\end{aligned}
\end{equation}
\begin{equation}\
\begin{aligned}
\bm V_h/\bm V_m=MLP(Norm(\Tilde{\bm V}^{s}_h/\Tilde{\bm V}^{s}_m))+\Tilde{\bm V}^{s}_h/\Tilde{\bm V}^{s}_m
\end{aligned}
\end{equation}
where $\bm V_h$/$\bm V_m$ denotes the output of a GLD block for input $\bm V^{s}_h$/$\bm V^{s}_m$. The [$CLS_h$] and [$CLS_m$] tokens output by the last block are used as the high-level global image feature $\bm v_{h}\in \mathbb{R}^{d}$ and middle-level global image feature $\bm v_{m}\in \mathbb{R}^{d}$ respectively. Besides, the image-level feature $\bm v_g$ that treats all local tokens in the same way is regarded as a low-level global feature $\bm v_l\in \mathbb{R}^{d}$. The above features constitute a multi-grained global image feature set $\bm V^g=\{\bm v_l, \bm v_m, \bm v_h\}$. Similarly, a similar process for text $T$ is performed to generate a multi-grained global text feature set $\bm T^g=\{\bm t_l, \bm t_m, \bm t_h\}$.

\begin{figure}[t!]
  \centering
  \includegraphics[width=3.4in,height=6.0in]{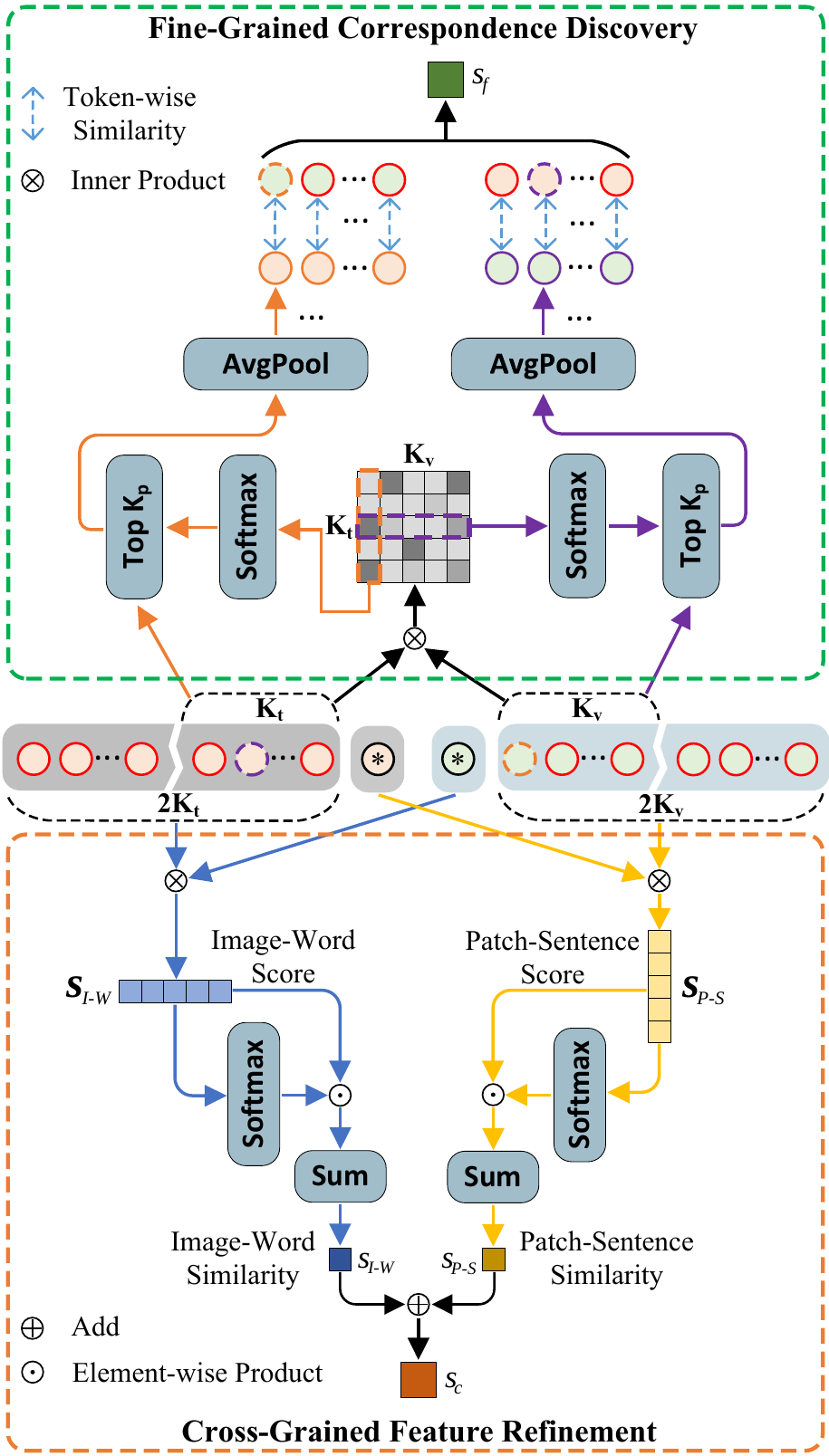}\\
  \caption{Illustration of cross-grained feature refinement (CFR) and fine-grained correspondence discovery (FCD) modules.} \label{Fig:4}
\end{figure}

\subsection{Cross-Grained Feature Refinement}
In MGF, we focus on mining intra-modal fine-grained discriminative information. The goal of TIReID is cross-modal alignment. Therefore, we hope that the mined information is not only highly discriminative but also modality-shared. Accordingly, we propose a cross-grained (i.e. image-word and text-patch) feature refinement (CFR) module to filter out unimportant and non-modality-shared information in the selected image and text tokens. Since token selection is based on the correlation between each local token and class token, we filter out unimportant information by the similarity between local and global features across modalities. The CFR is illustrated in Figure~\ref{Fig:4}(bottom).

Given the image-level representation $\bm v_{g}\in \mathbb{R}^{d}$, the selected patch-level representation $\bm V^{s}\in \mathbb{R}^{2K_{v}\times d}$, the sentence-level representation $\bm t_{g}\in \mathbb{R}^{d}$ and the selected word-level representation $\bm T^{s}\in \mathbb{R}^{2K_{t}\times d}$. The image-word and sentence-patch similarities are evaluated by inner products between corresponding feature representations, respectively.
\begin{equation}\
\begin{aligned}
\bm S_{I-W}=(\bm T^{s}\bm v_g)^T
\end{aligned}
\end{equation}
\begin{equation}\
\begin{aligned}
\bm S_{P-S}=\bm V^{s}\bm t_g
\end{aligned}
\end{equation}
where $\bm S_{I-W}\in \mathbb{R}^{1\times 2K_t}$ is the similarity between the image and the selected words in the sentence, and $\bm S_{P-S}\in \mathbb{R}^{2K_v\times 1}$  is the similarity between the sentence and the selected patches of an image. Then we fuse the above similarities to get the instance-level similarity. To emphasize important information and filter out non-modality-shared information in the selected image and text tokens, we adaptively generate different weights for each score in $\bm S_{I-W}$ and $\bm S_{P-S}$ by Softmax during aggregation, where scores for the image patches (words) related to the sentence (image) will be given high weights. Finally, the instance-level similarity is generated by computing a weighted sum of each score in $\bm S_{I-W}$ and $\bm S_{P-S}$. 
\begin{equation}\
\begin{aligned}
s_{I-W}=\sum_{i=1}^{2K_t}\frac{exp(\bm S_{I-W(1, i)})}{\sum_{j=1}^{2K_t}exp(\bm S_{I-W(1, j)})}\bm S_{I-W(1, i)}
\end{aligned}
\end{equation}
\begin{equation}\
\begin{aligned}
s_{P-S}=\sum_{i=1}^{2K_v}\frac{exp(\bm S_{P-S(i, 1)})}{\sum_{j=1}^{2K_v}exp(\bm S_{P-S(j, 1)})}\bm S_{P-S(i, 1)}
\end{aligned}
\end{equation}
\begin{equation}\
\begin{aligned}
s_{c}(I, T)=(s_{I-W} + s_{P-S}) / 2
\end{aligned}
\end{equation}
where $s_{c} \in\mathbb{R}^{1}$ denotes the similarity between image $I$ and text $T$ after cross-grained feature refinement. Since $I$ and $T$ are matched, we expect the greater the similarity the better. Through CFR module, the cross-modal correspondence between images and texts is roughly established. 

\subsection{Fine-Grained Correspondence Discovery}
To further mine fine-grained correspondence between images and texts, i.e., patch-word correspondence, we propose a fine-grained correspondence discovery (FCD) module, as shown in Figure~\ref{Fig:4}(top). The simplest way to establish the fine-grained correspondences is to find out the most related positive word (image patch) as the correspondence for each image patch (word) to construct the patch-word (word-patch) pairs, and make them close to each other in the feature space. However, individual word and patch may have vague meanings, and each word (patch) may be related to multiple patches (words), as shown in Figure~\ref{Fig:1}. Accordingly, we pick out the most related $K_p$ positive words (image patches) for each image patch (word) to construct the patch-word (word-patch) pairs. Moreover, to save computational costs, we only employ the most informative image patches and words for correspondence discovery.

Formally, for image-text pair $(I, T)$, the most informative patches and words are $\bm V^{s}_h\in \mathbb{R}^{K_{v}\times d}$ and $\bm T^{s}_h\in \mathbb{R}^{K_{t}\times d}$, respectively. We first compute the cosine similarity between $\bm V^{s}_h$ and $\bm T^{s}_h$. For example, the first row ${(\bm T^{s}_h{\bm V^{s}_h}^T)_{(1, :)}}\in \mathbb{R}^{K_{v}}$ of the similarity matrix denotes dependency of the $1^{th}$ word to each informative patch. Then,  the Topk operation is used to select the most related $K_p$ patches for the $1^{th}$ word $\bm t^{s}_1$. Finally, these $K_p$ patches are average pooled to generate a new patch feature $\bm v^{p}_1$, and $(\bm t^{s}_1, \bm v^{p}_1)$ forms a matched word-patch pair. 
\begin{equation}\
\begin{aligned}
~~~~~~~\bm v^{p}_1=AvgPool(Topk(Softmax((\bm T^{s}_h{\bm V^{s}_h}^T)_{(1, :)})))
\end{aligned}
\end{equation}

And so on, we can generate $K_v$ matched patch-word pairs $\{(\bm v^{s}_1, \bm t^{p}_1), ..., (\bm v^{s}_{K_v}, \bm t^{p}_{K_v})\}$ and $K_t$ matched word-patch pairs $\{(\bm t^{s}_1, \bm v^{p}_1), ..., (\bm t^{s}_{K_t}, \bm v^{p}_{K_t})\}$. To make these matched pairs close to each other in the feature space, we get the instance-level similarity by computing
the sum of the similarities between all $K_v+K_t$ matched pairs.
\begin{equation}\
\begin{aligned}
s_{f}(I, T)=\sum_{i=1}^{K_v}s(\bm v^{s}_i, \bm t^{p}_i)+\sum_{j=1}^{K_t}s(\bm v^{p}_j, \bm t^{s}_j)
\end{aligned}
\end{equation}
where $s(\cdot)$ denotes the cosine similarity metric, and $s_{f}$ denotes the similarity between image $I$ and text $T$ after fine-grained correspondence discovery. Similarly, the greater the similarity, the better.

\subsection{Training and Inference}
The commonly used cross-modal projection matching (CMPM) loss and cross-modal projection classification (CMPC) loss proposed by~\cite{CMPM} is adopted as our training objective function to learn image-text alignment. 

\textbf{Cross-Modality Alignment}. For image $I$ and text $T$, we generate multi-grained global feature sets $\{\bm v_l, \bm v_m, \bm v_h\}$ and $\{\bm t_l, \bm t_m, \bm t_h\}$ by MGF, respectively. To align $I$ and $T$, we compute CMPM and CMPC losses for each granularity of image and text features to supervise the multi-grained global feature learning.
\begin{equation}\
\begin{aligned}
\mathcal{L}_{cm}=\sum_{i\in[l,m,h]}(\mathcal{L}_{cmpm}^{i}+\mathcal{L}_{cmpc}^{i})
\end{aligned}
\end{equation}

Besides, since $s_{c}$ and $s_{f}$ both represent the similarity between $I$ and $T$, we sum them to represent the final similarity $s_{cf}$. For a batch of $B$ image-text pairs. We expect to maximize the similarity if the image and text are matched, and minimize it otherwise. Thus, a cross-modality bi-directional dual-constrained triplet ranking loss is used to optimize it.
\begin{equation}\
\begin{aligned}
\mathcal{L}_{c}&=max(\alpha-s_{cf}(I, T)+s_{cf}(I, T_n), 0)\\
&+max(\alpha-s_{cf}(I, T)+s_{cf}(I_n, T), 0)
\end{aligned}
\end{equation}
where $(I, T)$ denotes the matched image-text pairs, and $(I, T_n), (I_n, T)$ denote the mismatched pairs. $T_n$ and $I_n$ denote the hard negative samples. $\alpha$ indicates the margin.

\textbf{Diversity Regularization}. In order to fully mine fine-grained details in images and texts, we hope that different-grained features focus on inconsistent information. To this end, we impose a diversity constraint $\mathcal{L}_{div}$ on different-grained features to avoid information redundancy, which is represented as follows:
\begin{equation}\
\begin{aligned}
\mathcal{L}_{d}=\sum_{i\in[l,m,h]}\sum_{\mbox{\scriptsize$\begin{array}{c}j\in[l,m,h],\\ i\neq j\end{array}$}}(\frac{\bm v_i \cdot \bm v_j}{\|\bm v_i\|_2 \|\bm v_j\|_2}+\frac{\bm t_i \cdot \bm t_j}{\|\bm t_i\|_2 \|\bm t_j\|_2})
\end{aligned}
\end{equation}

\textbf{Objective Function}. Integrating the above constraints, the finial objection function $\mathcal{L}$ is as follows.
\begin{equation}\
\begin{aligned}
~~~~~~~\mathcal{L}=\mathcal{L}_{cm}+\lambda_c\mathcal{L}_{c}+\lambda_d\mathcal{L}_{d}
\end{aligned}
\end{equation}
where $\lambda_c$ and $\lambda_d$ balance the focus on different loss terms during training.

\textbf{Inference}. Note that CFR and FCD modules are only used for training, and will be removed to save computation costs for inference. During inference, for the text query and image candidate, we first generate the multi-grained global text and image features. Then, the similarity between text-image pair is computed as the sum of the cosine distance of different-grained features, i.e., $s=s_l+s_m+s_h$. 

\section{Experiments}
\subsection{Experiment Settings}
\subsubsection{Datasets and Metrics}
\textbf{CUHK-PEDES}~\cite{GNA} is previously the only accessible large-scale benchmark for TIReID. It includes 40,206 images and 80,412 text descriptions of 13,003 persons, each image is manually annotated with 2 descriptions, each of which has an average length of not less than 23 words. Follow~~\cite{GNA}, 34,054 images and 68,108 descriptions of 11,003 persons, 3,078 images and 6,156 descriptions of 1000 persons, 3,074 images and 6,148 descriptions of 1000 persons are utilized for training, validation, and testing, respectively. Recently, several large-scale datasets~\cite{SSAN, DSSL} have been released, which greatly promoted the development of TIReID. \textbf{ICFG-PEDES}~\cite{SSAN} contains 54522 text descriptions for 54,522 images of 4,102 persons collected from the MSMT17~\cite{MSMT17} dataset. According to statistics, each description has an average length of 37 words, and the vocabulary contains 5554 unique words. Compare with CUHK-PEDES, text description of ICFG-PEDES is more identity-centric and fine-grained. The dataset is split into train, and test with 34674 image-text pairs of 3102 persons, and 19848 image-text pairs of the remaining 1000 persons, respectively. \textbf{RSTPReid}~\cite{DSSL} is also constructed based on MSMT17~\cite{MSMT17} to handle real scenarios, which includes 41010 textual descriptions and 20505 images of 4101 persons. Specifically, each person contains 5 images caught by 15 cameras, each image corresponds to 2 text descriptions, and the length of each description is no shorter than 23 words. The dataset is split into 3701 train, 200 validation, and 200 test persons. We conduct sufficient experiments on the above three benchmarks to verify the effectiveness of our method. For all experiments, we use recall at Rank K (Rank-K, higher is better) as retrieval metric to evaluate the retrieval performance, where the Rank-1, Rank-5, and Rank-10 accuracy are reported.
\subsubsection{Implementation Details}
We conduct the experiments on a single RTX3090 24GB GPU using the PyTorch library. Input images are resized to 224$\times$224 and random horizontal flipping is employed for data augmentation. The input sentences are set with a maximum text length of 100 for all datasets. The length of visual and textual token sequences after tokenization is $N_v=196$ and $N_t=100$, and the dimension $d$ of image and text embeddings is 768. The multi-grained global feature learning module consists of $M=1$ GLD block, and every block has 12 heads. Besides, $2R_v=20\%$ informative image tokens and $2R_t=40\%$ informative text tokens in MGF are chosen to learn multi-grained global discriminative features. In the fine-grained correspondence discovery module, we pick out $K_p=3$ most related words (patches) for each patch (word) to construct the patch-word (word-patch) pair. The loss balance factors $\lambda_c$ and $\lambda_d$ are set to 10, 0.2, respectively. The margin $\alpha$ in triplet loss is set to 0.2. In the training process, we optimize our model with Adam optimizer and adopt a liner warmup strategy. We adopt different learning rates for different modules. To be specific, the initial learning rate for the image and text backbone is set to 1e-5, and other modules of the network are initialized to 1e-4. The learning rate is decreased by a factor of 0.1 at the 20th, 25th, and 35th epoch, respectively. The network is trained with a batch size of 32 and lasted for 50 epochs. 

\begin{table}[!ht]\small
\centering {\caption{Performance comparison with state-of-the-art methods on CUHK-PEDES. Rank-1, Rank-5, and Rank-10 are listed. '-' denotes that no reported result is available.}\label{Tab:1}
\renewcommand\arraystretch{1.2}
\begin{tabular}{c|c|ccc}
\hline
 \hline
  Methods  &Ref & Rank-1 & Rank-5 & Rank-10\\
  \hline

  GNA-RNN~\cite{GNA}  & CVPR17 & 19.05 & - & 53.64 \\
  
  GLA~\cite{GLA}   & ECCV18 & 43.58 & 66.93 & 76.26  \\

  Dual Path~\cite{Dual}   & TOMM20 & 44.40 & 66.26 & 75.07  \\

  CMPM/C~\cite{CMPM}  & ECCV18 & 49.37 & - & 79.27  \\

  MCCL~\cite{MCCL}  & ICASSP19 & 50.58 & - & 79.06  \\
  
  MIA~\cite{MIA}  & TIP20 & 53.10 & 75.00 & 82.90  \\

  A-GANet~\cite{A-GANet}  & MM19 & 53.14 & 74.03 & 81.95  \\
  
  PMA~\cite{PMA}  & AAAI20 & 53.81 & 73.54 & 81.23  \\

  TIMAM~\cite{TIMAM}  & ICCV19 & 54.51 & 77.56 & 84.78  \\

  CMKA~\cite{CMKA}   & TIP21 & 54.69  &73.65  &81.86 \\

  TDE~\cite{TDE}  & MM20 & 55.25 & 77.46 & 84.56  \\
  
  SCAN~\cite{SCAN}  & ECCV18 & 55.86 & 75.97 & 83.69 \\

  ViTAA~\cite{vitaa}  & ECCV20 & 55.97 & 75.84 & 83.52  \\

  IMG-Net~\cite{IMG-Net}  & JEI20 & 56.48 & 76.89 & 85.01  \\

  CMAAM~\cite{CMAAM}  & WACV20 & 56.68 & 77.18 & 84.86  \\

  HGAN~\cite{HGAN}  & MM20 & 59.00 & 79.49 & 86.62  \\

  SUM~\cite{sum}  & KBS22 & 59.22  & 80.35  & 87.60  \\

  NAFS~\cite{NAFS}   & arXiv21 & 59.94  &79.86  &86.70 \\

  DSSL~\cite{DSSL}   & MM21 & 59.98  &80.41  &87.56 \\
  
  MGEL~\cite{mgel}   & IJCAI21 & 60.27  &80.01  &86.74 \\

  SSAN~\cite{SSAN}   & arXiv21 & 61.37  &80.15  &86.73 \\
  
  LapsCore~\cite{LapsCore}  & ICCV21 & 63.40 & - & 87.80  \\
  
  IVT~\cite{ivt}   & ECCVW22 & 64.00  & 82.72   & 88.95 \\

  LBUL~\cite{LBUL} & MM22 & 64.04  & 82.66  & 87.22    \\
  
  TextReID~\cite{TextReID}   & BMVC21 & 64.08  &81.73  &88.19 \\
  
  SAF~\cite{saf}   & ICASSP22 & 64.13  &82.62  &88.40 \\

  TIPCB~\cite{tipcb}   & Neuro22 & 64.26  & 83.19  & 89.10 \\
  
  CAIBC~\cite{caibc}   & MM22 & 64.43  & 82.87   & 88.37 \\
  
  AXM-Net~\cite{AXM-Net}   & AAAI22 & 64.44  &80.52  &86.77 \\  
  \hline
  \textbf{Ours-IMG}  & - & \bf{65.07} & \bf{83.01} & \bf{89.00}  \\
  \hline
  \textbf{Ours-CLIP}  & - & \bf{69.57} & \bf{85.93} & \bf{91.15}  \\
  \hline\hline
\end{tabular}}
\end{table}

\subsection{Comparisons with State-of-the-art Models}
In this section, We evaluate our proposed CFine under two different settings, i.e., ViT pre-trained on ImageNet (Ours-IMG) or ViT from CLIP (Ours-CLIP) as image encoder, on three standard TIReID benchmarks and compare with state-of-the-art approaches in Tables~\ref{Tab:1}, ~\ref{Tab:2} and \ref{Tab:3}. Our method consistently achieves state-of-the-art results on all three benchmarks with significant improvements. 

\subsubsection{CUHK-PEDES}
We first evaluate our CFine on the most popular and widely-used benchmark \textbf{CUHK-PEDES}, and the performance comparison is shown in Table \ref{Tab:1}. It can be observed that CFine consistently achieves state-of-the-art results under two settings. With ViT pre-trained on ImageNet as image encoder, CFine reaches 65.07\%, 83.01\% and 89.00\% Rank-1, Rank-5, and Rank-10 accuracy on CUHK-PEDES, which outperforms the same Transformer-based methods, IVT~\cite{ivt} and SAF~\cite{saf}. In addition, CFine also achieves higher performance than the recent state-of-the-art method, named AXM-Net~\cite{AXM-Net}. This can be attributed to that our proposed fine-grained information excavation modules (MGF, CFR, and FCD) are critical to reducing the modality gap. Besides, AXM-Net requires cross-modality interaction operations, which are computationally expensive. With ViT from CLIP as image encoder, CFine can obtain significant gains in performance compared with all the methods. Compared with the strongest competitor (AXM-Net~\cite{AXM-Net}), CFine obtains 69.57\% (+5.13\%), 85.93\% (+5.41\%) and 91.15\% (+4.38\%) of Rank-1, Rank-5 and Rank-10 accuracy by employing ViT from CLIP as image encoder, which demonstrates that it is beneficial to introduce ample cross-modal correspondence prior.

\subsubsection{Other Benchmarks}
To further validate the generalization of our method, we also compare CFine against the previous works on two other benchmarks, \textbf{ICFG-PEDES} and \textbf{RSTPReid}, as shown in Tables~\ref{Tab:2} and~\ref{Tab:3}. We can observe from the tables that our method achieves very competitive performance on ICFG-PEDES and RSTPReid when using ViT pre-trained on ImageNet as image encoder, outperforming all methods except IVT~\cite{ivt}. While using ViT from CLIP as image encoder, the proposed method outperforms all existing methods by a large margin in all metrics, which reaches 60.83\%, 50.55\% Rank-1 accuracy on ICFG-PEDES and RSTPReid benchmarks respectively, surpassing the IVT~\cite{ivt} on Rank-1 by +4.79\%, +3.85\% respectively. Results on these benchmarks demonstrate the generalization and robustness of our proposed CFine.

\begin{table}[!ht]\small
\centering {\caption{Performance comparison with state-of-the-art methods on ICFG-PEDES. Rank-1, Rank-5, and Rank-10 are listed.}\label{Tab:2}
\renewcommand\arraystretch{1.2}
\begin{tabular}{c|c|ccc}
\hline
 \hline
  Methods & Ref & Rank-1 & Rank-5 & Rank-10 \\
  \hline
  Dual Path~\cite{Dual} & TOMM20 & 38.99 & 59.44 & 68.41 \\

  CMPM/C~\cite{CMPM} & ECCV18 & 43.51 & 65.44 & 74.26 \\

  MIA~\cite{MIA} & TIP20 & 46.49 & 67.14 & 75.18 \\

  SCAN~\cite{SCAN} & ECCV18 & 50.05 & 69.65 & 77.21 \\

  ViTAA~\cite{vitaa} & ECCV20 & 50.98 & 68.79 & 75.78 \\

  SSAN~\cite{SSAN} & arXiv21 & 54.23 & 72.63 & 79.53 \\

  TIPCB~\cite{tipcb}   & Neuro22  & 54.96  & 74.72   & 81.89 \\
  
  IVT~\cite{ivt}   & ECCVW22 & 56.04  & 73.60  & 80.22 \\
  \hline
  \textbf{Ours-IMG} & - & \bf{55.69} & \bf{72.72} & \bf{79.46} \\ 
  \hline
  \textbf{Ours-CLIP}  & - & \bf{60.83} & \bf{76.55} & \bf{82.42}  \\
  \hline\hline
\end{tabular}}
\end{table}

\begin{table}[!ht]\small
\centering {\caption{Performance comparison with state-of-the-art methods on RSTPReid. Rank-1, Rank-5, and Rank-10 are listed.}\label{Tab:3}
\renewcommand\arraystretch{1.2}
\begin{tabular}{c|c|ccc}
\hline
 \hline
  Methods & Ref & Rank-1 & Rank-5 & Rank-10 \\
  \hline
  IMG-Net~\cite{IMG-Net}  & JEI20 & 37.60 & 61.15 & 73.55  \\

  AMEN~\cite{amen}   & PRCV21  & 38.45  & 62.40  & 73.80 \\

  DSSL~\cite{DSSL}   & MM21 & 39.05  &62.60  &73.95 \\

  SUM~\cite{sum}   & KBS22 & 41.38  & 67.48  & 76.48 \\

  SSAN~\cite{SSAN} & arXiv21 & 43.50 & 67.80 & 77.15 \\

  LBUL~\cite{LBUL}  & MM22   & 45.55  & 68.20 & 77.85  \\
  
  IVT~\cite{ivt}   & ECCVW22 & 46.70  & 70.00  & 78.80 \\
  \hline
  \textbf{Ours-IMG} & - & \bf{45.85} & \bf{70.30} & \bf{78.40} \\
  \hline
  \textbf{Ours-CLIP}  & - & \bf{50.55} & \bf{72.50} & \bf{81.60}  \\
  \hline\hline
\end{tabular}}
\end{table}

The above results and analysis show that our CFine achieves consistent improvements across different benchmarks. This mainly attributes to the following factors. (1) our proposed fine-grained information excavation modules can fully mine intra-modal fine-grained discriminative details and inter-modal correspondences, effectively narrowing the modality gap and distinguishing different pedestrians. (2) Introducing ample cross-modal correspondences contained in CLIP can bring significant performance gains. (3) Benefiting from fine-grained information excavation, the cross-modal representation capacity of the upstream VLP task is successfully transferred to the TIReID task.

\begin{table*}[!ht]\small
\centering {\caption{Ablation study on different components of our proposed CFine on CUHK-PEDES.}\label{Tab:4}
\renewcommand\arraystretch{1.2}
\begin{tabular}{m{0.5cm}<{\centering}|m{3.2cm}<{\centering}|m{0.7cm}<{\centering}|m{0.7cm}<{\centering}|m{0.7cm}<{\centering}|m{0.7cm}<{\centering}|m{0.7cm}<{\centering}|m{0.7cm}<{\centering}|m{1.2cm}<{\centering}m{1.2cm}<{\centering}m{1.2cm}<{\centering}}
\hline
\hline
\multirow{2}{*}{No.} & \multirow{2}{*}{Methods} & \multirow{2}{*}{CLIP} &\multicolumn{3}{c|}{MGF}  & \multirow{2}{*}{CFR} & \multirow{2}{*}{FCD} & \multirow{2}{*}{Rank1} & \multirow{2}{*}{Rank5} & \multirow{2}{*}{Rank10}\\
\cline{4-6}
          &  &  & l & m & h  &  &  &       &       &  \\
\hline
0 & Baseline &  &\checkmark  &   &    &  &  & 57.89 & 78.46 & 85.77 \\

1 & +CLIP      &\checkmark &\checkmark &  &  &  &  & 65.56 & 83.74 & 89.85 \\

2 & +CLIP+MGF (m)     &\checkmark  &\checkmark &\checkmark  &  &  &  & 67.66 & 85.10 & 89.80 \\

3 & +CLIP+MGF (h)     &\checkmark  &\checkmark  &  &\checkmark  &  &  & 68.22 & 85.12 & 90.24 \\
  
4 & +CLIP+CFR  &\checkmark   &\checkmark   &    &  &\checkmark  &   & 66.41 & 84.16 & 89.91 \\
  
5 & +CLIP+FCD  &\checkmark  &\checkmark   &    &  &  &\checkmark   & 66.13 & 84.15 & 89.78 \\

6 & +CLIP+MGF  &\checkmark  &\checkmark &\checkmark  &\checkmark  &  &  & 68.62 & 85.36 & 90.84 \\

7 & +CLIP+MGF+CFR  &\checkmark  &\checkmark  &\checkmark  &\checkmark  &\checkmark  &  & 69.14 & 85.40 & 90.55 \\

8 & +CLIP+MGF+FCD  &\checkmark  &\checkmark  &\checkmark  &\checkmark  &  &\checkmark  & 69.33 & 85.22 & 90.61 \\

9 &     CFine      &\checkmark  &\checkmark   &\checkmark  &\checkmark  &\checkmark  &\checkmark & 69.57 & 85.93 & 91.15 \\
\hline\hline
\end{tabular}}
\end{table*}

\subsection{Ablation Studies}
To fully demonstrate the impact of different modules in CFine, we conduct extensive ablation studies to compare different variants of CFine on CUHK-PEDES. Here, we adopt ViT from CLIP as image encoder. To be specific, we first verify the contribution of each component in the model by combining different components. Important parameters and variants in each module are then discussed respectively.

\subsubsection{Contributions of Algorithmic Components}
We examine the contributions of each module in Table \ref{Tab:4}. No.0 shows the results of the Baseline. Baseline means only using ViT pre-trained on ImageNet and the language pre-trained model BERT as image and text encoders to extract features without adding any modules and further feature embedding. By comparing the results of No.0 and No.1 in Table \ref{Tab:4}, it can be observed that when replacing the image encoder of Baseline with ViT from CLIP, the performance increases by 7.67\%, 5.28\%, 4.08\% over Baseline on CUHK-PEDES under the Rank-1/5/10 accuracy and surpass all methods in Table \ref{Tab:1}, which shows that introducing cross-modal prior information into the model is able to bring significant performance gain. The results of No.1 vs No.2 and No.1 vs No.3 reveal the efficacy of MGF. When adding MGF (m) or MGF (h) to No.1, Rank-1 is improved by 2.1\% or 2.66\%, respectively. Besides, as shown in the result of No.6, the combination of MGF (m) and MGF (h) can further improve performance, which can promote the rank-1 accuracy from 65.56\% to 68.62\%. These results in No.2, No.3, and No.6 justify that MGF can effectively mine the discriminative local clues and learn discriminative global features at multiple granularities. 

CFR is used to filter redundant information and ensure confidence in informative tokens, while FCD is used to fully discover inter-modal fine-grained correspondences. These experimental results of No.1 vs No.4, No.1 vs No.5, No.6 vs No.7, and No.6 vs No.8 demonstrate the efficacy of CFR and FCD. When CFine is equipped with all modules, the best retrieval performance can be achieved. By comparing the results of No.1 and No.9, it can be proven that the direct usage of CLIP can be sub-optimal for TIReID due to the substantial gap between instance-level pre-training and fine-grained TIReID task. When combined with our proposed fine-grained information excavation modules (MGF, CFR, and FCD), the performance is improved from 65.56\% to 69.57\%. The result demonstrates that the ample cross-modal knowledge of CLIP is fully exploited and transferred to the TIReID task.

\subsubsection{Ablation of Multi-grained Global Feature Learning}
\begin{figure}[!t]
\centering
\setlength{\abovecaptionskip}{-5pt}
\includegraphics[width=3.5in,height=3.2in]{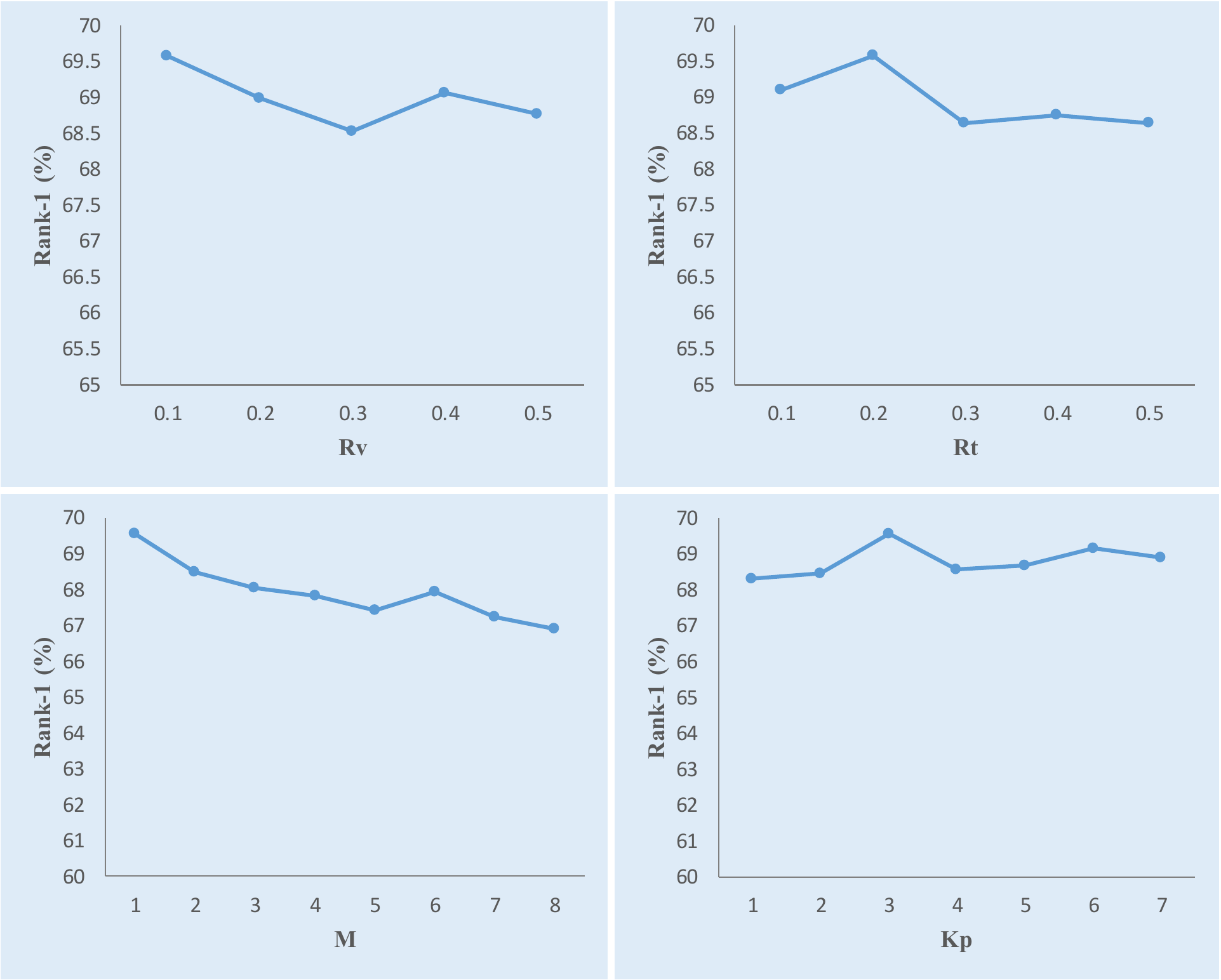}\\
\caption{Effect of different parameters, including (a) the selected ratio $R_v$ of image tokens (top left); (b) the selected ratio $R_t$ of text tokens (top right); (c) the amount $M$ of GLD blocks (bottom left); (d) the number $K_p$ of selected positive patches (words) for each word (patch) (bottom right).}
\label{Fig:5}
\end{figure}

In MGF, the token selection first selects the most informative image and text tokens, and then feeds them into GLD consisting of $M$ blocks to learn a set of multi-grained global features. We first analyze the impact of some important factors in MGF, including image token selection ratio $R_v$, text token selection ratio $R_t$, and amount $M$ of GLD blocks. Figure~\ref{Fig:5}(a)(b)(c) shows the impact of these parameters on performance. As shown in the figure, for $R_v$ and $R_t$, if the ratio is too small, it cannot contain comprehensive discriminative local information, resulting in information loss. If the ratio is too large, the noise will be introduced, which will have a negative impact on the model. Therefore, we set $R_v=0.1$ and $R_t=0.2$. For M, the results show that $M=1$ performs best.

\begin{table}[!ht]\small
\centering {\caption{Ablation study of multi-grained global feature learning on CUHK-PEDES.}\label{Tab:5}
\renewcommand\arraystretch{1.2}
\begin{tabular}{m{3cm}<{\centering}|m{1.2cm}<{\centering}m{1.2cm}<{\centering}m{1.2cm}<{\centering}}
\hline
 \hline
  Method  & Rank-1 & Rank-5 & Rank-10 \\
  \hline
  MGF-2K          & 67.43 & 84.80 & 90.27  \\

  Self-attention  & 68.14 & 84.76 & 90.09 \\

  $CLS$-mean        & 68.00 & 84.63 & 89.90  \\

  $CLS$-random      & 62.67 & 81.66 & 87.88  \\
  \hline
  Ours            & 69.57 & 85.93 & 91.15  \\
  \hline\hline
\end{tabular}}
\end{table}

We further conduct experiments to compare MGF with other variants, as shown in Table \ref{Tab:5}. In order to verify the effectiveness of the multi-grained learning way, we feed $2K_v/2K_t$ tokens into GLD at a time (MGF-2K) to learn discriminative global feature. The result shows severe performance degradation (-2.14\%). We believe that one-time learning way will cause the role of some important informative tokens to be overwritten. Through our multi-grained learning way, the discriminative fine-grained clues can be fully mined. In GLD, we conduct the cross-attention between selected tokens and all tokens to learn global features. Another possible choice is to compute the self-attention between selected tokens. Our scheme achieves more superior performance, which mainly attributes to our scheme can absorb information not only from informative tokens but also other context beyond them. In addition, before being fed into GLD, the high-level and middle-level token sequences are padded with [$CLS_h$] and [$CLS_m$] at the beginning to learn multi-grained global features, respectively. The initialization of these [$CLS$] tokens is critical. We compare three different initializations, including (1) initialization by the mean value of the selected token sequences ($CLS$-mean); (2) random initialization ($CLS$-random); (3) initialization by the global features output by the encoder (Ours). The results show that the third initialization achieves the best performance. We believe that initializing [$CLS$] with the output of encoder can strengthen the connection between GLD and encoder, make them cooperate with each other, and further emphasize the important fine-grained clues based on the output of encoder.

\begin{table}[!ht]\small
\centering {\caption{Effects of different similarity aggregation Scheme in cross-grained feature refinement module on CUHK-PEDES.}\label{Tab:6}
\renewcommand\arraystretch{1.2}
\begin{tabular}{m{3cm}<{\centering}|m{1.2cm}<{\centering}m{1.2cm}<{\centering}m{1.2cm}<{\centering}}
\hline
 \hline
  Method  & Rank-1 & Rank-5 & Rank-10 \\
  \hline
  Aggr-Sum   & 69.07  & 85.45   & 90.72  \\
  
  Aggr-Mean  & 68.66  & 85.32  &  90.46   \\

  Aggr-Max   & 68.44  & 85.15  &  90.11 \\
  \hline
  Ours  & 69.57 & 85.93 & 91.15   \\
  \hline\hline
\end{tabular}}
\end{table}

\subsubsection{Ablation of Cross-grained Feature Refinement}
In CFR, each score in image-word $\bm S_{I-W}$ and sentence-patch $\bm S_{P-S}$ similarities needs to be aggregated to form the instance-level similarity for the next text-image matching, in which score aggregation strategy is crucial. We compared different aggregation strategies, as shown in Table \ref{Tab:6}. For $2K_t$ ($2K_v$) scores in image-word (sentence-patch) similarity, we get the instance-level score by the following strategies: (1) Summing all scores (Aggr-Sum); (2) Taking the mean of all scores (Aggr-Mean); (3) Taking the maximum of these scores (Aggr-Max); (4) Computing a weighted sum of all scores, and the weights are generated by Softmax (Ours). The results show that our aggregation strategy achieves the best results, which can be attributed to the way of weighting each score can filter out the redundant information and highlight the important information.

\subsubsection{Ablation of Fine-grained Correspondence Discovery}
FCD picks out the most related words (patches) for each patch (word) for discovering fine-grained correspondence. We conduct experiments to verify what amount of $K_p$ is optimal for performance in Figure \ref{Fig:5}(d). The result shows that If $K_p$ is too small, the meaning will be ambiguous, and the network will pay too much attention to local correspondence, which will lead to over-fitting. However, if $K_p$ is too large, some irrelevant information will be introduced, making it impossible to establish an accurate correspondence. When $K_p=3$, our proposed method achieves the best results.

\begin{figure}[!t]
  \centering
  \includegraphics[width=3.5in,height=4.4in]{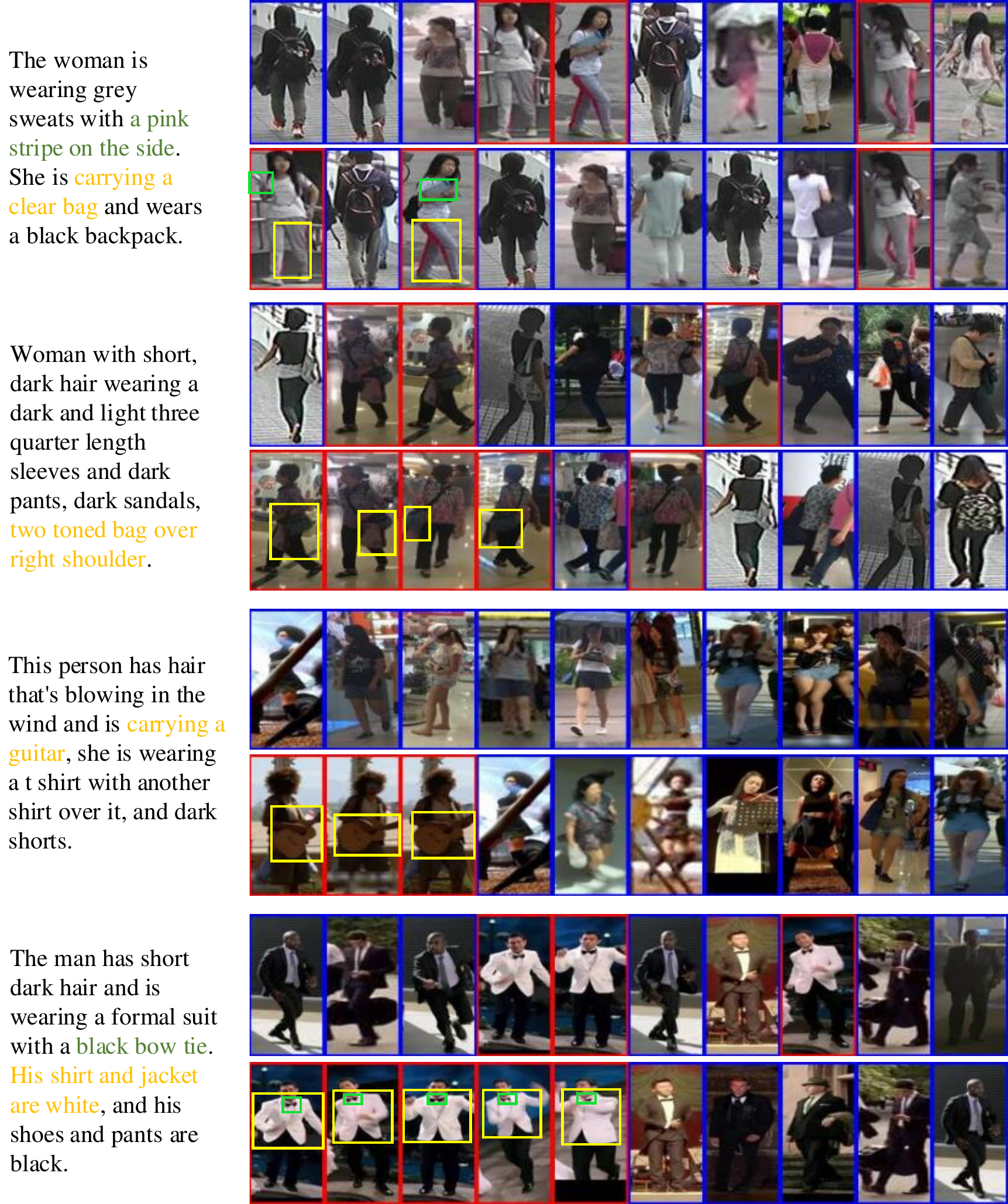}\\
  \caption{Comparison of top-10 retrieval results on CUHK-PEDES between Baseline+CLIP (the first row) and CFine (the second row) for each text query. The matched and mismatched person images are marked with red and blue rectangles, respectively.}\label{Fig:6}
\end{figure}

\subsubsection{Qualitative Results}
Figure \ref{Fig:6} shows the top-10 retrieval results of Baseline+CLIP and CFine for the given text query, the difference between the two is whether fine-grained information excavation is performed. It can be seen from the figure that CFine can still achieve accurate retrieval results in the case that some Baseline+CLIP fail to retrieve. This is mainly due to the fine-grained information excavation modules we designed, namely MGF, CFR, and FCD, which can fully mine the discriminative clues to distinguish different pedestrians. Besides, we also found an interesting phenomenon that by effectively mining local discriminative clues, such as bag, guitar, and shoe, the dependence of the model on color information can be reduced, which alleviates the color over-reliance problem of existing TIReID methods to a certain extent. This provides a new idea for us to further solve the color over-reliance problem in the future. 

\section{Conclusion}
In this work, we propose a CLIP-driven Fine-grained information excavation framework (CFine), a novel transformer architecture with fine-grained information excavation, which aims to leverage the power of CLIP to achieve cross-modal fine-grained alignment for TIReID. To take full advantage of the rich multi-modal knowledge from CLIP, we performed fine-grained information excavation. Specifically, MGF can help reinforce the intra-modal fine-grained discriminative clues by modeling the interactions between the global image (text) and local discriminative patches (words). By modeling the cross-grained interactions between modalities, CFR can filter out non-modality-shared local information and establish the rough cross-modal correspondence, followed by FCD to establish the fine-grained cross-modal correspondence by capturing the relationship between image patches and words. The above modules cooperate with each other to well transfer the knowledge of the CLIP model to TIReID. Additionally, we also show that feature embedding is not necessary for TIReID. Significant performance gains on three popular TIReID benchmarks prove the superiority and effectiveness of the proposed CFine.


%





\ifCLASSOPTIONcaptionsoff
  \newpage
\fi



%
\bibliography{mybibfile}
\end{document}